\documentclass[10pt,twocolumn]{article} 
\usepackage{simpleConference}
\usepackage{times}
\usepackage{graphicx}
\usepackage{amssymb}
\usepackage{url,hyperref}
\usepackage[T1]{fontenc}
\usepackage[utf8]{inputenc}
\usepackage{graphics}

\usepackage{amsmath}
\usepackage{pifont}
\usepackage{authblk}
\usepackage{pifont}
\usepackage{algorithm}
\usepackage{algpseudocode}
\usepackage{subcaption}
\usepackage{booktabs}
\usepackage{hyperref}
\usepackage{placeins}
\usepackage{listings}
\begin{document}

\title{Multiobjective Hydropower Reservoir Operation Optimization with Transformer-Based Deep Reinforcement Learning}
\author[a]{Rixin Wu}
\author[a]{Ran Wang \thanks{Corresponding author}}
\author[a]{Jie Hao}
\author[a]{Qiang Wu}
\author[b]{Ping Wang}
\affil[a]{College of Computer Science and Technology, Nanjing University of Aeronautics and Astronautics, Nanjing, China}
\affil[b]{Deptartment of Electrical Engineering and Computer Science, Lassonde School of Engineering, York University, Canada}
\renewcommand*{\Affilfont}{\small\it} 
\renewcommand\Authands{ and } 
\date{} 

\maketitle
\thispagestyle{empty}

\begin{abstract}
	Due to shortage of water resources and increasing water demands, the joint operation of multireservoir systems for balancing power generation, ecological protection, and the residential water supply has become a critical issue in hydropower management. However, the numerous constraints and nonlinearity of multiple reservoirs make solving this problem time-consuming. To address this challenge, a deep reinforcement learning approach that incorporates a transformer framework is proposed. The multihead attention mechanism of the encoder effectively extracts information from reservoirs and residential areas, and the multireservoir attention network of the decoder generates suitable operational decisions. The proposed method is applied to Lake Mead and Lake Powell in the Colorado River Basin. The experimental results demonstrate that the transformer-based deep reinforcement learning approach can produce appropriate operational outcomes. Compared to a state-of-the-art method, the operation strategies produced by the proposed approach generate 10.11\% more electricity, reduce the amended annual proportional flow deviation by 39.69\%, and increase water supply revenue by 4.10\%. Consequently, the proposed approach offers an effective method for the multiobjective operation of multihydropower reservoir systems.
\end{abstract}

\section{Introduction}\label{sec:sec1}
~\

As a clean and renewable resource that generates no pollution, hydropower is being extensively developed \cite{xu2015adaptive} in response to the growing strain on the Earth's traditional energy sources \cite{gao2022optimal,di2018water}. The conventional hydropower operation scheme typically focuses on determining the optimal water level or power generation capacity for all reservoirs to maximize overall economic benefits. This operation model is straightforward to implement and has experienced success in real-world applications \cite{kumar2023effective}. Regrettably, the economic advantages of reservoirs often come at the expense of the natural ecological health of rivers \cite{ma2013short}. A high flow rate is often maintained for power generation, and electricity is over generated at the cost of disrupting the downstream environment. This imbalance ultimately leads to ecological degradation \cite{madani2009modeling}. Simultaneously, in real-world operations, hydropower reservoirs must serve multiple purposes, such as supplying domestic, industrial, and irrigation water \cite{yao2022long}. Based on these factors, the coordinated operation of multiple hydropower reservoirs is needed.

Multiobjective multihydropower reservoir operation optimization (MMROO) has emerged as a vital and complex task in modern hydropower reservoir systems \cite{niu2018parallel}. As the duration of reservoir operation increases, particularly when dealing with numerous reservoirs and many areas requiring water, both the scale of the problem and the challenge of resolving it intensify. The number of decision variables is positively correlated with the number of reservoirs, water supply area, number of planning years, and inverse of the time step \cite{niu2021multiple,ma2021spark}. Considering multiple objectives, such as power generation, environmental protection, and water supply benefits, further complicates the operational system. Consequently, traditional hydropower reservoir management approaches struggle to meet people's needs. As a result, developing a practical multihydropower reservoir operation model and an efficient algorithm for the model has become a pressing concern \cite{yao2022long}.

In this paper, we innovatively develop an MMROO model that balances power generation, ecological protection, and water supply benefits. To address the MMROO problem, we utilize a transformer-based deep reinforcement learning approach. The main contributions of this paper are summarized as follows:
\begin{itemize}
	
	\item In terms of the system model, we propose a multihydropower reservoir model tailored to meet practical needs. Specifically, a single reservoir often cannot meet the supplying water needs for agricultural irrigation, industry, and domestic use. Accordingly, a multireservoir coordinated operation approach is better suited to address real-world requirements.
	
	\item In terms of problem formulation, we develop a multiobjective optimization model to address diverse requirements in hydropower reservoir operation. This model comprehensively considers the maximization of power generation and water supply benefits as well as the minimization of the amended annual proportional flow deviation (AAPFD) value \footnote{The AAPFD value can measure the ecological stability of the river, and the smaller the AAPFD value, the more stable the river ecology.}.
	
	\item In terms of algorithms for solving the MMROO problem, we devise a transformer-based deep reinforcement learning (T-DRL) method and adopt a two-stage encoder process for information embedding. This approach provides higher solution efficiency than direct deep reinforcement learning method, as well as superior generalization ability and adaptability compared to the most commonly used multi-objective evolutionary algorithms: non-dominated sorting genetic algorithm-III (NSGA-III) and difference-based multiobjective evolutionary algorithm (MOEA/D). The proposed operation strategy not only enhances the power generation schemes of hydropower reservoirs but also guarantees a higher level of ecological protection, thus providing a well-rounded approach to reservoir management.
	
	\item In terms of the experimental results, our algorithm demonstrates excellent ability to produce effective operation strategies. When compared to that obtained with a state-of-the-art method, the operational strategy produced by the proposed approach generates 10.11\% more electricity, decreases the AAPFD value by 39.69\%, and increases water supply revenue by 4.10\%. These outcomes highlight the effectiveness and advantages of our method in managing hydropower reservoirs.
	
\end{itemize}

The remainder of this paper is organized as follows. Related work is introduced in Section \ref{sec:sec2}. In Section \ref{sec:sec3}, we present the system model for hydropower reservoir operation, along with the objective functions and constraints within the MMROO model. In Section \ref{sec:sec4}, the details of the T-DRL method for solving the MMROO problem are introduced. In Section \ref{sec:sec5}, we present a regional case study and analyze the results of model implementation. Finally, we conclude our paper in Section \ref{sec:sec6}.

\section{Related work}\label{sec:sec2}
~\
The operation of hydropower reservoirs focuses on the efficient allocation of water resources to accommodate various needs, such as power generation, residential water supply, and agricultural irrigation. The operation process must account for various physical conditions, including reservoir runoff and inflow. This is a classic problem within hydropower systems. In terms of system modeling, the problem may involve single-reservoir operation or multireservoir operation. In multireservoir operation scenarios, several reservoirs often need to collaborate to accomplish specific operational tasks. With respect to operation objectives, the problem may involve single-objective optimization or multiobjective optimization.

In the early stages of operation optimization for hydropower reservoirs, single-objective optimal operation methods for single reservoirs were often applied. Researchers have proposed a variety of methods to address the single-objective optimal operation problem for individual reservoirs. Ju-Hwan Yoo applied a linear programming model to the Yongtan multipurpose dam in Jinjiang, South Korea, to maximize hydropower production, resulting in a 184 GWh increase in energy production \cite{YOO2009182}. However, linear programming methods are difficult to apply to nonlinear systems. In \cite{saadat2017reliability}, reliability-improved stochastic dynamic programming (RISDP) was employed to ensure that the reservoir storage capacity approached the optimal value. Utilizing the RISDP operation strategy improved the objective function value by approximately 15\% compared to that in the actual case and eliminated the need for line conditions. Nevertheless, dynamic programming, as a type of nonlinear method, faces challenges in problems with high-dimensional datasets. Evolutionary algorithms are widely employed to optimize hydropower reservoir operation due to their high efficiency in solving complex problems (high-dimensional, nonconvex, and discrete issues). Among various evolutionary algorithms, genetic algorithms (GAs) are the most prevalent \cite{vasan2009comparative,chong2021optimization}. The authors of \cite{vasan2009comparative} compared simulated annealing (SA), simulated quenching (SQ), and a GA with the aim of maximizing the annual net benefits of irrigation planning. The results indicated that all three algorithms could be effectively used to meet irrigation demand and scheduling objectives. In \cite{chong2021optimization}, a parameter-free Jaya algorithm was utilized to minimize the total deficit of hydropower production, proving more effective than a GA, the ant colony algorithm (ACO), and several other existing algorithms. Single-reservoir single-objective operation optimization often involves simple systems and objectives, while actual reservoir systems tend to be complex.

As the operational demands of hydropower reservoirs have increased, single-objective optimization models have become insufficient. Consequently, some researchers have proposed hydropower reservoir operation strategies based on single-reservoir multiobjective optimization. In single-reservoir multiobjective operation problems, the most commonly considered objectives are power generation and ecological protection \cite{HE2020124919,yu2021multi,wu2021use}. He et al. conducted a multiobjective optimization of the operation of a large deep reservoir with the goals of maximizing total power generation, minimizing the root mean square errors of inflow and outflow, and maximizing the ecological index, and the nondominated genetic algorithm-II (NSGA-II) was applied to solve the problem \cite{HE2020124919}. In another study, a multiobjective game theory model (MOGM) was applied to balance economic, social, and ecological benefits in the operation of the Three Gorges Reservoir \cite{yu2021multi}. The progressive optimality algorithm-particle swarm optimization (POA-PSO) method in \cite{wu2021use} was used to harmonize power generation, environmental impacts, and water supply needs. Moreover, the maximization of hydropower generation and the minimization of the water supply deficit were simultaneously optimized in \cite{wang2021comparison}. While the single-reservoir multiobjective reservoir operation strategy considers multiple objectives for simultaneous optimization, it is essential to recognize that in complex systems, multiple reservoirs often need to collaborate to complete intricate tasks.

The multireservoir multiobjective operation strategy considers a more universally applicable system model in which multiple reservoirs are jointly dispatched to fulfill diverse demands. Guo et al. optimized the operation of multireservoir systems to maximize the lowest water level and the number of periods, using the improved nondominated particle swarm optimization (I-NSPSO) algorithm to solve the problem \cite{guo2013multi}. The authors of \cite{niu2018parallel} employed parallel multiobjective particle swarm optimization (MOPSO) to optimize the generation benefits of cascade hydropower reservoirs and the stable power output of hydropower systems. Accounting for the actual function of reservoirs, some studies consider flood control, domestic water supply, and agricultural water supply as optimization objectives \cite{hatamkhani2019multi,feng2018optimization}. Multireservoir multiobjective operation optimization is the most prevalent method in practical systems, as it can satisfy all system requirements. Currently, multiobjective evolutionary algorithms, such as NSGA-II and MOPSO, are primarily used to solve multiobjective optimization models. However, regarding the joint operation of multiple reservoirs, the speed and accuracy of conventional algorithms may not be satisfactory, especially when the system experiences disturbances. In such cases, evolutionary algorithms must be optimized entirely \cite{leng2022multi}.

\renewcommand\tablename{\textrm{Table}}
\begin{table*}[ht]
	\centering
	\caption{\textrm{Summary of existing methods for hydropower reservoir operation}}
	\begin{tabular}{c c c c c l}
		\hline
		\textrm{Reference} & \textrm{Method} & \textrm{System model} & \textrm{Power generation} & \textrm{Ecological protection} & \textrm{Other}\\ \hline
		\textrm{\cite{YOO2009182}},\textrm{\cite{CHONG2021107325}} & \textrm{LP, Jaya} & \textrm{single-reservoir} & \ding{51} & &\\ \hline
		\textrm{\cite{saadat2017reliability}} & \textrm{RISDP} & \textrm{single-reservoir} & & & \textrm{storage capacity}\\ \hline
		\textrm{\cite{vasan2009comparative}} & \textrm{SA, SQ, GA} & \textrm{single-reservoir} & & & \textrm{annual net benefits}\\ \hline
		\textrm{\cite{HE2020124919}} & \textrm{NSGA-II} & \textrm{single-reservoir} & \ding{51} & \ding{51} &\\ \hline
		\textrm{\cite{yu2021multi}} & \textrm{MOGM} & \textrm{single-reservoir} & \ding{51} & \ding{51} & \textrm{social objective}\\ \hline
		\textrm{\cite{wu2021use}} & \textrm{POA-PSO} & \textrm{single-reservoir} & \ding{51} & \ding{51} & \textrm{water supply}\\ \hline
		\textrm{\cite{wang2021comparison}} & \textrm{Differential evolution} & \textrm{single-reservoir} & \ding{51} & & \textrm{water supply}\\ \hline
		\textrm{\cite{guo2013multi}} & \textrm{I-NSPSO} & \textrm{multi-reservoir} & & & \textrm{water level, time periods}\\ \hline
		\textrm{\cite{niu2018parallel}} & \textrm{Parallel MOPSO} & \textrm{multi-reservoir} & \ding{51} & \ding{51} &\\ \hline
		\textrm{\cite{hatamkhani2019multi}} & \textrm{MOPSO} & \textrm{multi-reservoir} & \ding{51} & & \textrm{agricultural development}\\ \hline
		\textrm{\cite{xu2020deep}} & \textrm{DRL} & \textrm{single-reservoir} & \ding{51} & &\\ \hline
	\end{tabular}
	\label{table:summary}
\end{table*}

With the advancement of artificial intelligence technology, methods based on machine learning have been proposed to tackle optimization problems. As a subfield of machine learning, reinforcement learning (RL) serves as a data-driven approach that requires fewer system details and effectively addresses relevant problems. Over the past few decades, RL has been extensively applied in various domains, including path planning \cite{liu2020integrating}, network resource allocation \cite{cui2019multi}, and planning and scheduling optimization \cite{kedir2022hybridization}. However, the applications of RL techniques in water resource and hydropower systems are scarce \cite{xu2021deep}. Meanwhile, as the scale of the problem has expanded, RL methods have struggled to efficiently solve large-scale problems with various combinations of states and actions, resulting in the curse of dimensionality issue \cite{mnih2021playing}.

Recently, deep reinforcement learning (DRL) techniques have evolved by combining traditional RL with deep learning representations of nonlinear, high-dimensional mappings between system states and expected action rewards \cite{Volodymyr2015Human,zhang2023multi}. In the recent literature, DRL techniques have also been applied for the operational optimization of hydropower reservoirs. In \cite{SUN2020115660}, the authors trained a deep Q-learning network (DQN) agent to manage optimal storage reservoirs. Xu et al. developed a DRL framework based on a newly defined knowledge sample form and a DQN \cite{xu2020deep}. They used an aggregation-disaggregation model to reduce the dimensionality of the reservoir and employed three DRL models to realize the intelligent operation of cascade reservoirs. Although DRL technology has been developed for many years, its application in hydropower reservoir scheduling is still limited, particularly in multiobjective cases. A comprehensive overview of existing hydropower reservoir operation schemes can be found in Table \ref{table:summary}.

In our study, we apply a transformer-based deep reinforcement learning (T-DRL) approach to solve the MMROO problem. Previous multiobjective optimization studies \cite{niu2018parallel, HE2020124919, wang2021comparison, guo2013multi, feng2018optimization} did not account for various functions, such as power generation and ecological protection, nor did they consider the scenario of multireservoir joint operation. In our work, we propose a three-objective optimization model based on power generation, ecological protection, and water supply benefits. This model can appropriately describe the scenario of multireservoir joint operation.

	\section{Problem statement}\label{sec:sec3}
~\
\renewcommand\tablename{\textrm{Table}}
\begin{table}[ht]
	\centering \caption{\textrm{Nomenclature used in this paper}}
	\begin{tabular}{l p{6cm}}
		\hline 
		\textrm{Symbol} & \textrm{Definition}  \\
		\hline
		$I$& \textrm{Set of hydropower reservoirs}  \\
		$J$ & \textrm{Set of residential areas}  \\
		$T$ & \textrm{Set of operation periods}  \\
		$\Delta t$ & \textrm{Time interval in period $t$}  \\
		${A_i}$ & \textrm{Power coefficient of reservoir $i$} \\
		$V_i^{beg}$ &\textrm{Initial storage of reservoir $i$}  \\
		$V_{i,t}$ & \textrm{Storage volume of reservoir $i$ in period $t$}  \\
		$L_{i,t}$ & \textrm{Elevation of reservoir $i$ in period $t$}  \\
		$P_{i,t}$ & \textrm{Power generation of reservoir $i$ in period $t$}  \\
		$Q_{i,t}^e$ & \textrm{The most ecologically suitable runoff of reservoir $i$ in period $t$}\\
		$Q_{i,t}^p$ & \textrm{Turbine discharge of reservoir $i$ in period $t$}\\
		$Q_{i,t}^r$& \textrm{Inflow of reservoir $i$ in period $t$}\\
		$Q_{i,j,t}^s$ & \textrm{Flow of supplied water from reservoir $i$ to residential area $j$ in period $t$}\\
		$L_{i,t}^{min}$ $L_{i,t}^{max}$& \textrm{Minimum and maximum elevations of reservoir $i$ in period $t$}\\
		$P_{i,t}^{min}$ $P_{i,t}^{max}$& \textrm{Minimum and maximum power generation levels of reservoir $i$ in period $t$}\\
		$W_{j,t}^{min}$ $W_{j,t}^{max}$& \textrm{Minimum and maximum water supply volumes for residential area $j$ in period $t$}\\
		$A A P F D_{i}$ & \textrm{AAPFD value of reservoir $i$}\\
		$H_{i,t}$ & \textrm{Water head of reservoir $i$ in period $t$}\\
		$l_{i,j}$ & \textrm{Distance between reservoir $i$ and residential area $j$}\\	
		$b_{j,t}$ & \textrm{Unit water benefit for residential area $j$ in period $t$}\\
		$B_{i,j,t}$ & \textrm{Revenue from reservoir $i$ for residential area $j$ in period $t$}\\
		$x_{i,j,t}$ & \textrm{State vector of $Q_{i,j,t}^s$ (binary)}  \\
		$c_{i,j,t}$ & \textrm{Unit cost of water supply from reservoir $i$ for residential area $j$ in period $t$}\\
		${d_i}\left(\cdot  \right)$ & \textrm{Nonlinear function between storage and the elevation of reservoir $i$}\\
		\hline
	\end{tabular}
	\label{nomenclature}
\end{table}

\begin{figure}
	\centering
	\includegraphics[scale=1.1]{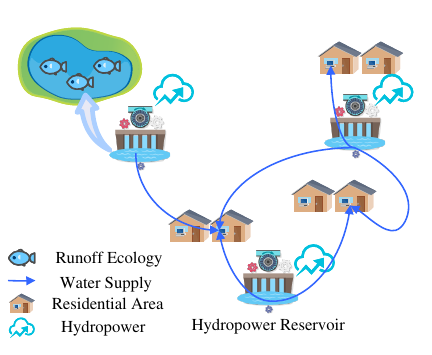}
	\caption{\textrm{An illustration of a multihydropower reservoir system}}
	\label{system}
\end{figure}

In this section, a formal description of MMROO is introduced. As depicted in Figure \ref{system}, a network of multiple geodistributed hydropower reservoirs is established to generate electricity while simultaneously supplying water to several residential areas. However, fulfilling these needs can result in adverse impacts on downstream ecosystems due to hydropower reservoir operation. To address this issue and achieve a balance between ecological concerns and reservoir functionality, we incorporate ecological requirements into reservoir operation. The primary nomenclature utilized throughout this paper, along with the corresponding meanings, is presented in Table \ref{nomenclature}. In the following subsections, we provide a detailed description of the system model and problem formulation.

	\subsection{System model}\label{sec:sec31}
~\
\subsubsection{Power generation}\label{sec:sec311}
We divide the operational period into time slots of the same length. Let $I$ denote the set of hydropower reservoirs. We further denote the turbine discharge of reservoir $i$ in period $t$ as $Q_{i,t}^{p}$, the water head of reservoir $i$ in period $t$ as ${H_{i,t}}$, the power coefficient of reservoir $i$ as ${A_i}$, and the duration of period $t$ as $\Delta t$. With these definitions in place, the total power generation of reservoir $i$ in period $t$ is defined as follows:

\begin{equation} \label{eq:power}
	{P_{i,t}} = {A_i}Q_{i,t}^{p}{H_{i,t}}{\Delta t}.
\end{equation}

\subsubsection{Ecological protection}\label{sec:sec312}
In the process of hydropower reservoir operation, ecological protection encompasses two primary aspects: river ecology and vegetation ecology \cite{tan2021identifying}. For river ecology, runoff ecology refers to the amount of water required to maintain the ecological function of the river, provided that certain water quality standards are met. The most suitable ecological flow supports the spawning, survival, and reproduction of indicative species, thereby ensuring the stability and integrity of the river ecosystem. When the flow is significantly lower than the most suitable ecological level, the river water quality may deteriorate, and the river may dry up or even disappear \cite{he2019impact,jia2020ecological}. Conversely, if the flow substantially exceeds the suitable ecological level, flooding, soil submersion, and swamping can occur \cite{fang2020multi}.

The Amended Annual Proportional Flow Deviation (AAPFD) was shown to effectively reflect the health of river ecosystems in previous studies \cite{feng2019adapting,liu2019multi}. A small AAPFD value indicates a healthy river ecology. We further define the ecological flow of reservoir $i$ in period $t$ as $Q_{i,t}^e$. Under such a definition, the AAPFD value of reservoir $i$ during the entire operation period can be defined as follows:

\begin{equation} \label{eq:AAPFD}
	A A P F D_{i}=\sqrt{\sum_{t=1}^T\left(\frac{Q_{i, t}^p-Q_{i, t}^e}{Q_{i, t}^e}\right)^2}.
\end{equation}

\subsubsection{Water supply}\label{sec:sec313}
Considering the practical applications of hydropower reservoirs, in our system model, the reservoirs are designed to supply water to nearby residential areas. Let $J$ denote the set of residential areas. Considering the varying distances between different reservoirs and residential areas, the costs of supplying unit water from reservoirs to residences may differ significantly. As such, for the same residential area, the decision on whether or not to supply water from different reservoirs, and the respective quantities supplied, can influence one another. It's worth noting that water supply to a residential area isn't restricted to a single reservoir, and multiple reservoirs may contribute to the water supply simultaneously. 

Therefore, we define a binary variable ${x_{i,j,t}} = 0/1$ to indicate whether water is delivered from reservoir $i$ to residential area $j$ in period $t$ or not. The unit water income for residential area $j$ in period $t$ is denoted as $ b_{j,t}$. We define the cost of supplying a unit of water from reservoir $i$ to residential area $j$ in period $t$ as $c_{i,j,t}$, the flow required to supply water from reservoir $i$ to residential area $j$ in period $t$ as $Q_{i,j,t}^s$, and the distance between reservoir $i$ and residential area $j$ as $l_{i,j}$. With these definitions, the total revenue produced by reservoir $i$ for residential area $j$ in period $t$ can be expressed as follows:

\begin{equation} \label{eq:water}
	B_{i,j,t}=\left[ {{b_{j,t}}Q_{i,j,t}^s - {c_{i,j,t}}{l_{i,j}}Q_{i,j,t}^s} \right]{x_{i,j,t}}\Delta t.
\end{equation}

\subsection{Problem formulation}\label{sec:sec32}
~\
In this section, we provide a detailed description of the three objective functions and physical constraints in the MMROO problem. Given water resource limitations, the aim of MMROO is to simultaneously achieve the maximization of power generation, the minimization of the ecological AAPFD value, and the maximization of water supply benefits.
\subsubsection{Decision variables}\label{sec:sec323}
The MMROO problem involves the following decision variables:

$Q_{i,t}^{p}$: the power generation flow from reservoir $i$ in period $t$;

${x_{i,j,t}}$: whether water is delivered from reservoir $i$ to residential area $j$ in period $t$ or not;

$Q_{i,j,t}^s$: water supply flow from reservoir $i$ to residential area $j$ in period $t$.

\subsubsection{Objective functions}\label{sec:sec321}

	1. Maximizing total power generation
	
	The primary purpose of designed hydropower reservoirs is to convert potential water-based energy into electrical energy \cite{niu2018parallel, zhang2022ultra}. Hence, the first objective function we select in the MMROO problem is to maximize the total power generation of all hydropower reservoirs during operation periods, which can be expressed as follows:

	\begin{equation} \label{eq:obj1}
		F_{power}=\max \sum_{i=1}^I \sum_{t=1}^T P_{i, t}.
	\end{equation}
	2. Minimizing the ecological AAPFD value
	
	Considering the sustainable development of river ecology, some hydropower reservoirs have environmental requirements \cite{nicklow2010state}. As introduced in Section \ref{sec:sec312}, the AAPFD value reflects the health of a river, with a healthy river ecology exhibiting a low AAPFD value. Therefore, the objective function of minimizing the AAPFD value can be represented as follows:

	\begin{equation} \label{eq:obj2}
		F_{A A P F D}=\min \sum_{i=1}^I A A P F D_{i}.
	\end{equation}
	3. Maximizing the total water supply benefit
	
	In the practical application of hydropower reservoirs, some reservoirs are required to supply water to nearby residential areas. When dealing with multireservoir joint operations, the distance between each reservoir and each residential area must be considered in the model. As a result, the third objective function is to maximize the total water supply benefit, which can be expressed as follows:

	\begin{equation} \label{eq:obj3}
		F_{water} = \max \sum\limits_{i = 1}^I {\sum\limits_{j = 1}^J {\sum\limits_{t = 1}^T{B_{i,j,t}}}}.
	\end{equation}

\subsubsection{Constraints}\label{sec:sec322}

\begin{itemize}
	\item [(a)]
	Water balance constraints:
	\begin{equation}\label{eq:con1}
		\begin{array}{r}
			V_{i, t}=V_{i, t-1}+\left[Q_{i, t}^r-Q_{i, t}^p-\sum_{j=1}^J Q_{i, j, t}^s x_{i, j, t}\right] \Delta t, \\
			i \in[1, I], t \in[1, T].
		\end{array}
	\end{equation}
	
	\item[(b)]
	Water elevation constraints:
	\begin{equation}\label{eq:con2}
		L_{i,t}^{\min } \le {L_{i,t}} \le L_{i,t}^{\max },i \in \left[ {1,I} \right],t \in \left[ {1,T} \right].
	\end{equation}
	
	\item [(c)]
	Power generation constraints:
	\begin{equation}\label{eq:con3}
		P_{i,t}^{\min } \le {P_{i,t}} \le P_{i,t}^{\max },i \in \left[ {1,I} \right],t \in \left[ {1,T} \right].
	\end{equation}
	
	\item [(d)]
	Water supply constraints:
	\begin{equation}\label{eq:con4}
		\begin{array}{r}
			W_{j,t}^{\min } \le \sum\limits_{i = 1}^I {Q_{i,j,t}^s{x_{i,j,t}}}{\Delta t} \le W_{j,t}^{\max },\\
			j \in \left[ {1,J} \right],t \in \left[ {1,T} \right].
		\end{array}
	\end{equation}
	
	\item [(e)]
	Initial condition constraints:
	\begin{equation}\label{eq:con5}
		{V_{i,0}} = V_i^{beg},i \in \left[ {1,I} \right].
	\end{equation}
	
	\item [(f)]
	Nonlinear relationship constraints:
	\begin{equation}\label{eq:con6}
		{L_{i,t}} = d_i\left( {{V_{i,t}}} \right),i \in \left[ {1,I} \right],t \in \left[ {1,T} \right].
	\end{equation}
	
\end{itemize}

In this model, constraint (\ref{eq:con1})  calculates the storage volume of each reservoir in each period according to the inflow flow, power generation flow and water supply flow. Constraint (\ref{eq:con2}) ensures that the elevation of the reservoir is within the specified range. Constraints (\ref{eq:con3}) and (\ref{eq:con4}) limit on power generation and water supply. Constraint (\ref{eq:con5}) guarantees the initial storage volume of the reservoir. Constraint (\ref{eq:con6}) defines the nonlinear relationship between reservoir elevation and storage volume.

	\section{Methodology}\label{sec:sec4}
~\

Given the complexity of the MMROO problem, the existing reservoir operation methods appear to be inadequate for effectively addressing various issues. Therefore, in this section, we introduce a transformer-based deep reinforcement learning (T-DRL) approach to solve the proposed MMROO problem. We begin by outlining the general framework of T-DRL, and a detailed explanation of the decomposition strategy employed to solve the MMROO problem is then given. Next, we discuss the transformer architecture, specifically the encoder and decoder processes. Finally, we provide a description of the training process.

\subsection{General framework}\label{sec:sec41}
~\

In the MMROO problem, a wide range of information pertaining to reservoirs and residential areas, such as maximum and minimum power generation and water supply, must be considered. As a result, specialized information extraction techniques are required to effectively process these high-dimensional data. Shallow or simple neural networks are evidently incapable of processing the detailed information required in MMROO. However, the transformer architecture, which employs attention mechanisms, has been proven to excel in tasks such as sequence modeling and machine translation within the natural language processing (NLP) domain \cite{vaswani2017attention, 9354025}. Furthermore, recent research has explored the integration of transformer architectures with DRL methods for solving optimization problems, demonstrating superior performance compared to traditional methods \cite{9847118,9978654}.

\begin{figure}
	\centering
	\includegraphics[scale=0.75]{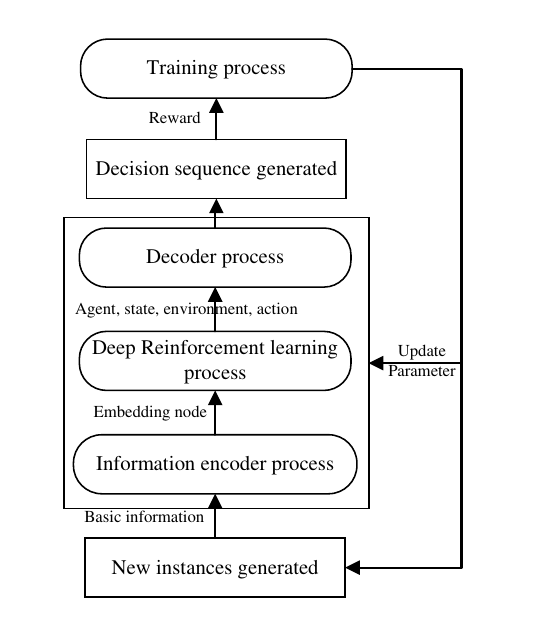}
	\caption{\textrm{Framework of the transformer-based DRL method}}
	\label{framework}
\end{figure}

\begin{figure}[ht]
	\centering
	\begin{subfigure}[b]{0.28\textwidth}
		\includegraphics[width=\textwidth]{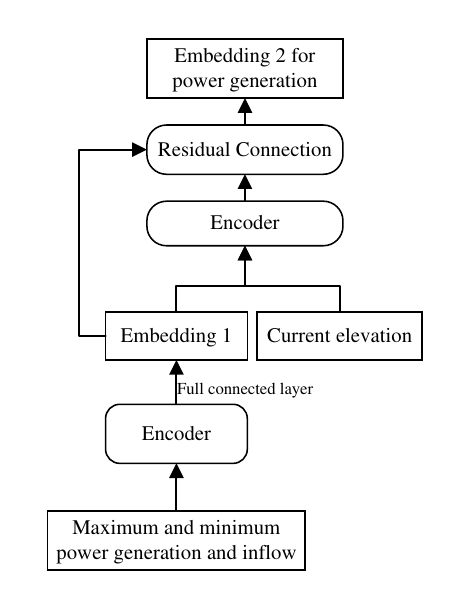}
		\caption{Two-stage embedding}
		\label{emd1a}
	\end{subfigure}
	\hfill
	\begin{subfigure}[b]{0.19\textwidth}
		\includegraphics[width=\textwidth]{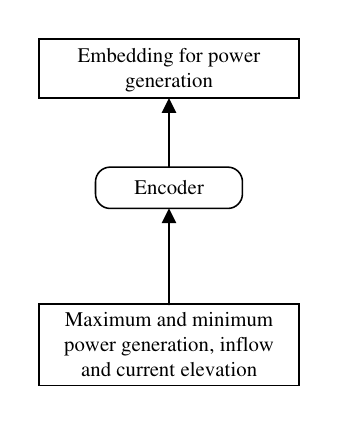}
		\caption{Direct embedding}
		\label{emd1b}
	\end{subfigure}
	\caption{\textrm{The process of Embedding for power generation. (a) involves a two-stage learning progress, while (b) inputs all information directly to the encoder.}}
	\label{emd1}
\end{figure}

\begin{figure}[ht]
	\centering
	\begin{subfigure}[b]{0.28\textwidth}
		\includegraphics[width=\textwidth]{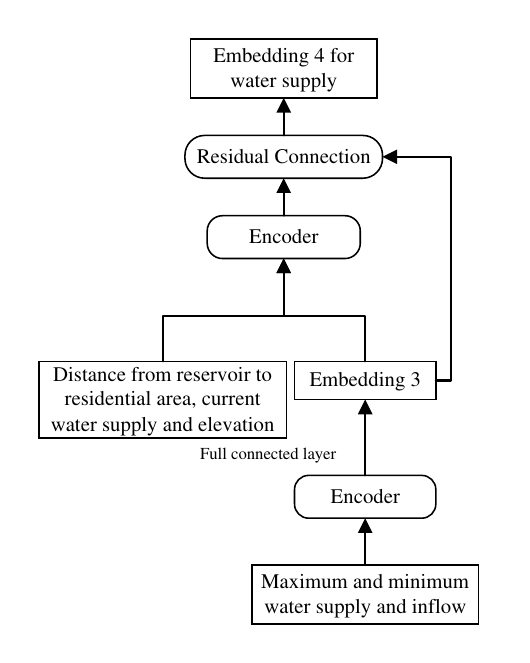}
		\caption{Two-stage embedding}
		\label{emd2a}
	\end{subfigure}
	\hfill
	\begin{subfigure}[b]{0.19\textwidth}
		\includegraphics[width=\textwidth]{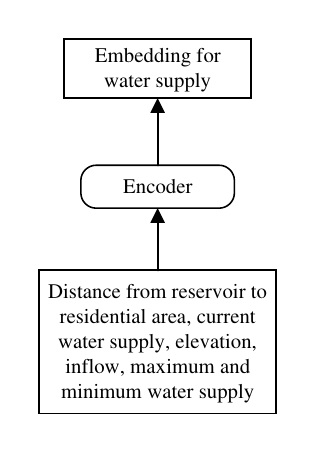}
		\caption{Direct embedding}
		\label{emd2b}
	\end{subfigure}
	\caption{\textrm{The process of Embedding for water supply. (a) involves a two-stage learning progress, while (b) inputs all information directly to the encoder.}}
	\label{emd2}
\end{figure}

As depicted in Figure \ref{framework}, our method is divided into three main parts: the encoder process, deep reinforcement learning process, and decoder process. During each training iteration, newly generated epoch instances are fed into the transformer architecture. The primary objective of the encoder process is to generate embeddings for power generation via multiple reservoirs and for water supply to multiple residential areas. The reservoir embedding process accounts for the monthly maximum and minimum power generation as well as the average inflow information. In contrast, the residential area embedding process primarily involves the maximum and minimum monthly water supplies. On this basis, the deep reinforcement learning process and decoder process are employed to generate the sequence of decision variables. During this phase, we provide detailed definitions for agents, actions, environments, and rewards. The multihead attention layer is used to generate reservoir operation decisions during the decoder process. Ultimately, the gradients obtained from the reward are backpropagated to optimize the parameters of the neural network. The parameters are trained jointly in an end-to-end fashion.
\renewcommand{\algorithmicrequire}{\textbf{Input:}}
\renewcommand{\algorithmicensure}{\textbf{Output:}}

\begin{algorithm}
	\caption{The decoder process of the transformer model in subproblem $a$}
	\begin{algorithmic}[1]
		\Require The reservoir embedding $x_{i, t}^{(1)}$, the residential area embedding $x_{i, j, t}^{(2)}$, and the initial elevation $L_{i, t}$ and inflow $Q_{i, t}^r$;
		\Ensure The operation decision and reward $R_a$
		
		\State \textbf{Begin}
		\For{$t = 1$ to $T$}, 
		\For{$i = 1$ to $I$},
		\State Compute Embedding 2 based on $x_{i, t}^{(1)}$ and $L_{i, t}$;
		\State Choose the action $Q_{i, t}^p$ for power generation;
		\State Update the current elevation $L_{i,t}$ by $Q_{i, t}^p$ and $Q_{i, t}^r$;
		\For{$j = 1$ to $J$},
		\If{$i == 1$}, \Comment{no water supply from other reservoir before}
		\State Set the current water supply $W_{j, t}=0$;
		\Else
		\State Set $W_{j, t}=\sum_{m=1}^{i-1} Q_{m, j, t}^s x_{m, j, t} \Delta t$; \Comment{by previous water supply from other reservoirs}
		\EndIf
		\State Compute Embedding 4 based on $x_{j, t}^{(2)}$, $L_{i, t}$, $l_{i, j}$ and $W_{j, t}$;
		\State Choose action $x_{i, j, t}$ for whether to supply water;
		\If{$x_{i, j, t} !=0$},
		\State Choose action $Q_{i, j, t}^s$ for the water supply;
		\Else
		\State  Set $Q_{i, j, t}^s=0$;
		\EndIf
		\State Update the current elevation $L_{i,t}$ by $Q_{i, j, t}^s$;
		
		\EndFor
		\EndFor
		\EndFor
		\If{The operation decision satisfies Eqs. (\ref{eq:con2})-(\ref{eq:con4})},
		\State Set $R_a$ as Eq. (\ref{eq:reward})
		\Else
		\State Set $R_a=0$;
		\EndIf
		\State \textbf{End}
	\end{algorithmic}
	\label{al1}
\end{algorithm}
\subsection{Decomposition strategy}\label{sec:sec42}
~\
Multiobjective optimization problems (MOPs) are commonly decomposed into sets of standardized optimization problems using the widely adopted linear weighting method. Solving this set of standardized optimization problems yields the Pareto front of the MOP \cite{wang2023robust}. We break down the MMROO problem, which comprises three objective functions, into 171 subproblems through weight combination with a mutual interval of 0.05:$w_{a, b}=[[0.05,0.05,0.9],[0.05,0.1,0.85], ...,[0.9,0.05,0.05]]$, where $w_{a,b}$ represents the weight of objective function $b$ in subproblem $a$. This particular weighting combination can ensure that the resulting Pareto front displays both considerable adaptability and a relatively even distribution of solutions. For each subproblem, the objective function, which is also related to the reward in deep reinforcement learning, can be determined through the three objective functions and their corresponding weights. Simultaneously, since the three objective functions in this study have different dimensions, directly summing the weighted objective function values and weights would result in a Pareto-optimal solution that is biased toward the objective function with a larger dimension. To address this issue, we employ the max-min normalization method to map the objective function values to the interval [0,1]. Additionally, considering that
the second objective function seeks to minimize the AAPFD value, the reward function $R_a$ for subproblem $a$ is defined as follows:
\begin{equation}\label{eq:reward}
	\begin{aligned} R_a= & w_{a, 1} \frac{F_{a, power}-F_{power }^{\min }}{F_{power}^{\max }-F_{power }^{\min }}+\\ &w_{a, 2} \frac{1 / F_{a, A A P F D}-1 / F_{A A P F D}^{\max }}{1 / F_{A A P F D}^{\min }-1 / F_{A A P F D}^{\max }}+\\ &w_{a, 3} \frac{F_{a,  water }-F_{water }^{\min }}{F_{water }^{\max }-F_{power}^{\min }},\end{aligned}
\end{equation}
where $F_{power}^{\max }$, $F_{ power }^{\min }$, $F_{ A A P F D} ^{\max }$, $F_{ A A P F D}^{\min }$, $F_{ water}^{\max }$ and $F_{ water }^{\min }$ represent the maximum and minimum values of the three objective functions, respectively. All of these values are obtained through single-objective T-DRL. In subproblem a, $F_{a, power}$, $F_{a, AAPFD}$, and $F_{a, water}$ denote the values of the three corresponding objective functions.  By evaluating the three objective functions across all subproblems, we can derive the Pareto front for the MMROO problem.

	\subsection{Encoder in the transformer model}\label{sec:sec43}
~\
Compared to single-reservoir single-objective operation problems, the MMROO problem encompasses not only power generation from multiple reservoirs but also water supply to residential areas. Consequently, processing this information simultaneously is not feasible due to the distinct differences among the corresponding datasets. Therefore, a critical challenge in the encoder design process is the integration of both reservoir information and residential area information.

The MMROO problem involves diverse and distinct decision variables related to power generation and water supply, which requires the implementation of multiple encoders to effectively process the information. For the generation of power generation decisions, the information that needs to be considered in the whole process includes the maximum and minimum power supply and the elevation, which are two different types of information. Traditional encoding method often input them directly into the neural network, but this approach can compromise stability during the learning phase. We therefore develop a two-stage learning strategy to better learn different types of information.

Figure \ref{emd1} illustrates the embedding framework for power generation information. Figure \ref{emd1a} represents the two-stage embedding process (denoted as Two-stage T-DRL) with two embedding layers responsible for general reservoir information ($Q_{i,t}^{\min}$, $Q_{i,t}^{\min}$ and $Q_{i,t}^r$) and the current water level ($L_{i,t}$). Figure \ref{emd1b} inputs the information above into the transformer architecture directly (denoted as Direct T-DRL). Figure \ref{emd2} displays the embedding framework for water supply information. Similar to the above, \ref{emd2a} employs the two-stage T-DRL method to generate the embedding for the water supply decision, while \ref{emd2b} utilizes the Direct T-DRL method for the same purpose.

The initial Embedding 1 for reservoir $j$, which corresponds to the general reservoir information embedding $x_{i,t}$ \cite{zhang2022transformer, perera2023graph}, is obtained using the following formula:
\begin{equation}\label{eq:embedding}
	x_{i, t}=W_1\left[P_{i, t}^{\min }, P_{i, t}^{\max }, Q_{i, t}^r\right]+b_1, i \in[1, I], t \in[1, T],
\end{equation}
where the operation $[\cdot, \cdot ,\cdot]$ concatenates three tensors of the same dimension. Subsequently, the multihead attention layer is employed to process the embedding $x_{i,t}$ and map it to a key $k_{i,t}$, query $q_{i,t}$, and value $v_{ i,t}$. The output $x_{i, t}^{(1)}$ of the self-attention layer is calculated by weighting the value $v_{i,u}$ by normalized dot product between the query $q_{i,t}$ and other keys $k_{i,u}$:
\begin{equation}\label{eq:xout}
	\begin{array}{r}
		x_{i, t}^{(1)}=\sum_{u=1}^T \operatorname{softmax}\left[\left\{q_{i,t}, k_{i,u^{\prime}}\right\}_{u^{\prime}=1}^T\right]_u v_{i,u},\\i \in[1, I], t \in[1, T].
	\end{array}
\end{equation}

Through the above calculation process, the encoder outputs $x_{i, t}^{(1)}$ for power generation and the encoder outputs $x_{i, j, t}^{(2)}$ for water supply are respectively calculated.

\subsection{Decoder of the transformer model}\label{sec:sec44}
~\
We model the decoder process as a Markov decision process, consisting of the agents (each reservoir), the state set $S$, the action set, which includes $A^{p}$ for power generation, $A^{x}$ for deciding whether to supply water, and $A^{s}$ for supplying water to residential areas, the reward function $R$ and the observed environment set $E$.

For each hydropower reservoir $i$, the operation decision-making process is as follows. In every period $t$, the environmental state $e_t \in E$ is determined, and a power generation water decision $Q_{i, t}^p \in A^p$ is produced. Subsequently, $L_{i, t}$ is updated to acquire a new state, and water supply operation decisions $x_{i, j, t} \in A^x$ and $Q_{i, j, t}^s \in A^w$ are made. This process is carried out for each residential area.

The purpose of the agent is to learn a policy through repeated learning to maximize the reward function, as defined in Eq.(\ref{eq:reward}). A summary of the decoder process is presented in Algorithm \ref{al1}.

\subsection{Training process}\label{sec:sec45}
~\
The policy gradient method with baseline \cite{sutton1999policy} is applied to our neural network to train the parameters $\theta$. First, the advantage estimation function of subproblem $a$ is determined based on the following equation:
\begin{equation}\label{eq:ADV}
	A D V_{a,i}=R_{a}\left(\pi_{a,i}\right)-R_{a}\left(\pi_a^{B L}\right),
\end{equation}
where $\pi_{a,i}$ represents the policy generated by the proposed method in subproblem $a$, $R_{a}\left(\pi_a^{B L}\right)$ represents the reward obtained with the baseline model in subproblem $a$. Next, the parameters are updated via:
\begin{equation}\label{eq:para}
	\nabla_\theta L_a(\theta)=\frac{1}{B} \sum_{i=1}^B A D V_{a,i} \nabla_\theta \log p_\theta\left(\pi_{a,i}\right),
\end{equation}
where$\nabla_\theta \log p_\theta(\pi_{a,i})$ represents the gradient of the logarithm of the probability distribution with respect to the model parameters $\theta$ in subproblem $a$. $B$ represents the batch size. Throughout the training process, a paired t test is conducted to compare $\theta$ and $\theta^{B L}$. If the result is found to be significant at the 95\% confidence level, $\theta^{B L}$ is replaced by $\theta$. This step ensures that the updated parameters provide a statistically significant improvement over the previous parameters, thereby refining the model's performance.
\begin{figure}
	\centering
	\includegraphics[scale=0.65]{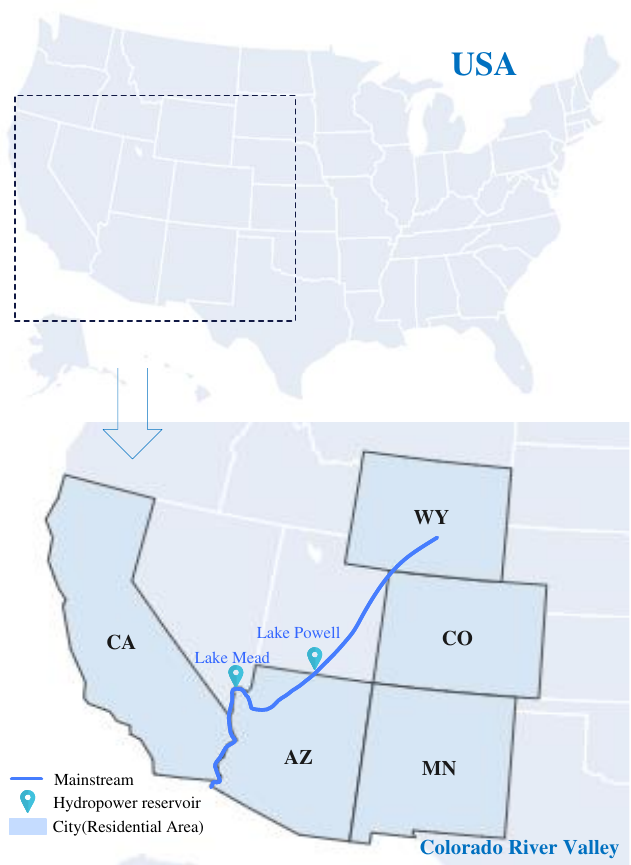}
	\caption{\textrm{Brief view of Lake Mead and Lake Powell}}
	\label{view}
\end{figure}
\section{Case study}\label{sec:sec5}
~\
In this section, the proposed method is applied to determine the optimal operation plan for a dual-hydropower reservoir system in the Colorado River Valley.

	\subsection{Study area}\label{sec:sec51}
~\

In this study, we focus on two key hydropower reservoirs, namely, the Glen Canyon Dam at Lake Powell and the Hoover Dam at Lake Mead, to validate the effectiveness of our proposed model. As illustrated in Figure 5, both Lake Powell and Lake Mead play crucial roles in supplying water to five states in the United States: Arizona (AZ), California (CA), Wyoming (WY), New Mexico (NM), and Colorado (CO).

According to the Colorado River Basin August 2022 24-Month Study released by the Bureau of Reclamation, the region has been experiencing prolonged drought and low-runoff conditions, exacerbated by climate change, leading to historically low water levels in both Lake Powell and Lake Mead \cite{ColoradoSurvey}. Over the past two decades, authorities have collaborated with Colorado River Basin partners to implement various drought response measures. Despite these efforts, water levels continue to decrease, emphasizing the need for efficient utilization of the limited water resources available.

The Glen Canyon Dam, located 15 miles upstream of Lees Ferry, serves as the primary feature of the Colorado River Storage Project (CRSP). Boasting more storage capacity than all other facilities of the CRSP combined, the Glen Canyon Dam plays a crucial role in the water and power resource management of the upper Colorado River Basin.

Situated in the Black Canyon of the Colorado River, approximately 35 miles southeast of Las Vegas, Nevada, the Hoover Dam and Lake Mead straddle the Arizona-Nevada state line.
\begin{table}
	\centering
	\caption{\textrm{Parameters of T-DRL}}
	\begin{tabular}{cc}
		\toprule
		\textrm{Description} & \textrm{Value} \\
		\midrule
		\textrm{Platform} &  \textrm{Pytorch 1.11}\\
		\textrm{Learning rate} & \textrm{1e-3(epoch<3)}\\       \textrm{} &\textrm{1e-4(epoch$\geq$3)\cite{wang2019distributed}}\\
		
		\textrm{Batch size} & \textrm{128}\\
		\textrm{Embedding size} & \textrm{128}\\
		\textrm{Number of attention heads} & \textrm{8}\\
		\textrm{Maximum number of epochs} & \textrm{5}\\
		\textrm{Iteration number per epoch} & \textrm{200}\\
		\textrm{Optimizer} & \textrm{Adam\cite{zhang2022transformer}}\\
		
		\bottomrule
	\end{tabular}
	\label{parameter}
\end{table}

\begin{figure*}[ht]
	\scriptsize
	\begin{tabular}{cc}
		\includegraphics[width=8.5cm]{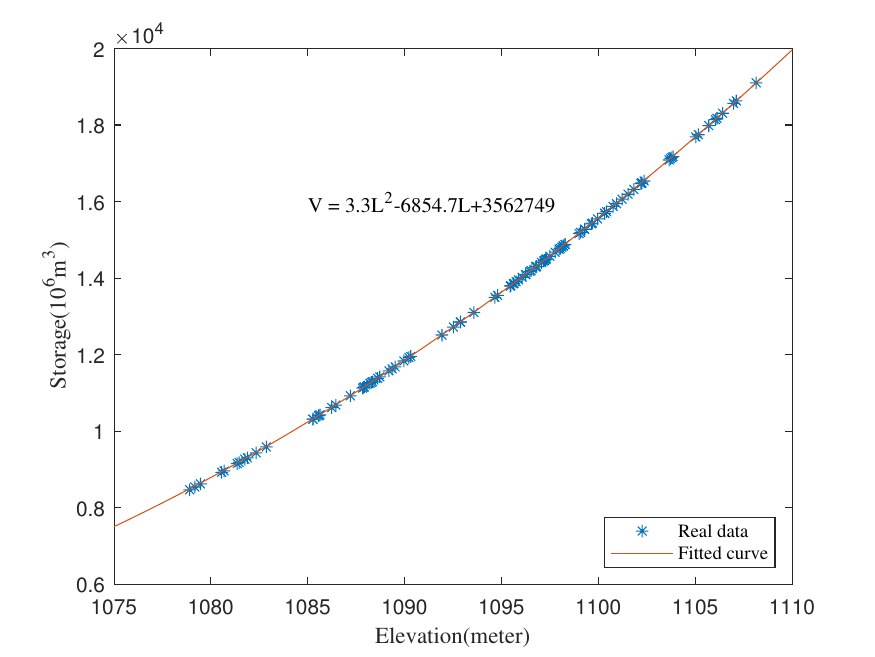} &    \includegraphics[width=8.5cm]{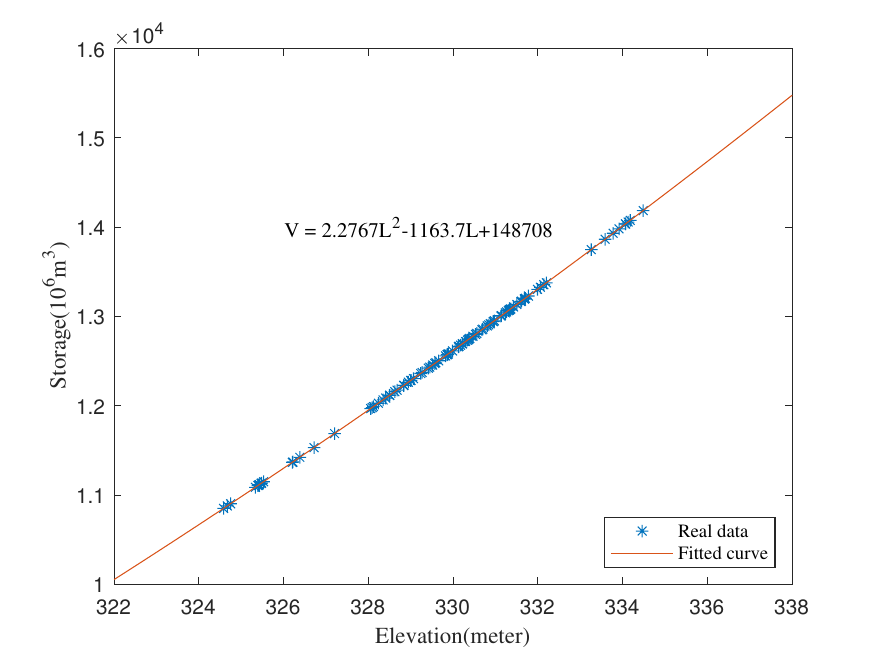}        \\
		\textrm{Lake Powell Glen Canyon Dam} & \textrm{Lake Mead Hoover Dam} \\
	\end{tabular}
	\caption{\textrm{Nonlinear relationship between elevation and storage for the two hydropower reservoirs}}
	\vspace{-0.5em}
	\label{elevation}
\end{figure*}
\begin{table}[]
	\setlength{\tabcolsep}{7pt}
	
	\caption{\textrm{The most suitable
			ecological outflow
			for the two reservoirs (unit: $m^3/s$).}}
	\begin{tabular}{ccc}
		\toprule
		\textrm{{Month}} & {\textrm{Lake Powell}} & {\textrm{Lake Mead}} \\
		\midrule
		\textrm{January} & \textrm{314.3869} & \textrm{259.0656} \\
		\textrm{February} & \textrm{265.4993} & \textrm{243.4002} \\
		\textrm{March} & \textrm{259.9415} & \textrm{293.6558} \\
		\textrm{April} & \textrm{341.7421} & \textrm{337.1620} \\
		\textrm{May} & \textrm{247.8343} & \textrm{229.2298} \\
		\textrm{June} & \textrm{242.0624} & \textrm{296.4402} \\
		\textrm{July} & \textrm{223.1525} & \textrm{240.8004} \\
		\textrm{August} & \textrm{353.4624} & \textrm{395.0267} \\
		\textrm{September} & \textrm{381.5739} & \textrm{354.9071} \\
		\textrm{October} & \textrm{332.6873} & \textrm{288.5897} \\
		\textrm{November} & \textrm{248.5995} & \textrm{211.9601} \\
		\textrm{December} & \textrm{249.7109} & \textrm{187.3235} \\
		\bottomrule
	\end{tabular}
	\label{MSER}
\end{table}
\subsection{Parameter setting}\label{sec:sec52}
~\
\subsubsection{Parameters in the algorithm}\label{sec:sec521}
To assess the performance of our proposed Two-stage T-DRL approach in solving the MMROO problem, we compare it to three widely used multiobjective optimization algorithms: the nondominated sorting genetic algorithm-III (NSGA-III), the multiobjective evolutionary algorithm based on decomposition (MOEA/D), and Direct T-DRL. The parameters for each of these algorithms are detailed below.

\begin{itemize}
	\item
	The parameters for NSGA-III are as follows: the population size is set to 200; the mutation probability is 10\%; the crossover probability is 90\%; the coding type is "real encoding"; and the maximum generations is set to 100.
	
	\item
	The parameters for MOEA/D are as follows: the population size is set to 200; the neighborhood size is 20; the maximum number of generations is 100; the update probability is 50\%; the mutation probability is 10\%; and the crossover probability is 90\%.
	
	\item
	The parameters for Two-stage T-DRL and Direct T-DRL are presented in Table \ref{parameter}.
\end{itemize}

\subsubsection{Parameters in the model}\label{sec:sec522}
The parameters in the model, including the basic settings for Lake Powell and Lake Mead, are outlined below. The majority of these parameters are obtained from the U.S. Bureau of Reclamation website \cite{usbr_data}. The nonlinear relationship between water elevation and storage volume for both reservoirs is depicted in Figure \ref{elevation}.
\begin{figure*}[!htb]
	\centering
	\subfloat[View of the three coordinate axes\label{PF1}]{\includegraphics[width=7cm]{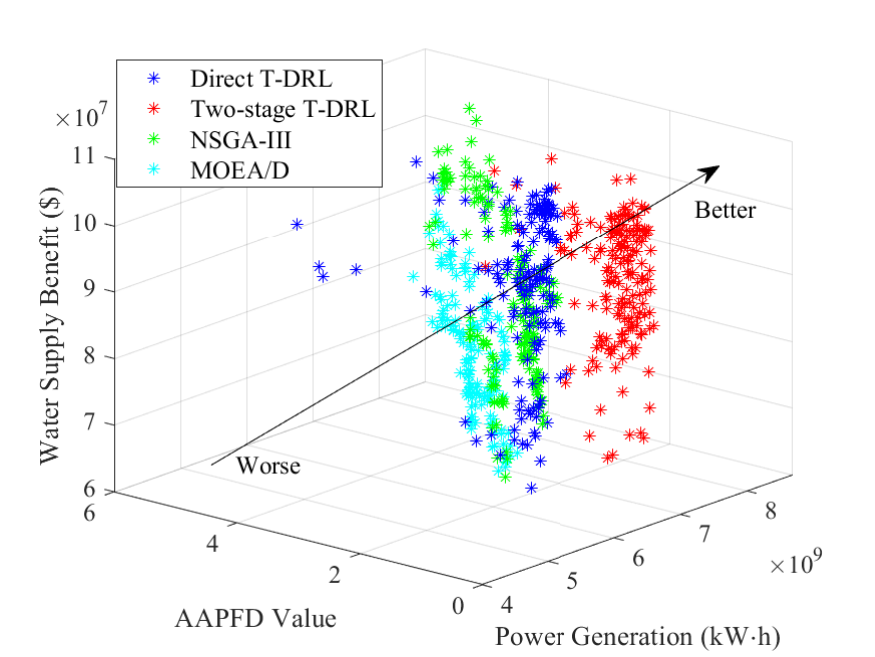}}
	\subfloat[View of the X and Z axes\label{PF2}]{\includegraphics[width=7cm]{PF1.pdf}}
	
	\subfloat[View of the Y and Z axes\label{PF3}]{\includegraphics[width=7cm]{PF1.pdf}}
	\subfloat[View of the X and Y axes\label{PF4}]{\includegraphics[width=7cm]{PF1.pdf}}
	\caption{Pareto front comparison of the four methods}
	\label{PF}
\end{figure*}

\begin{figure}[ht]
	\centering
	\includegraphics[scale=0.6]{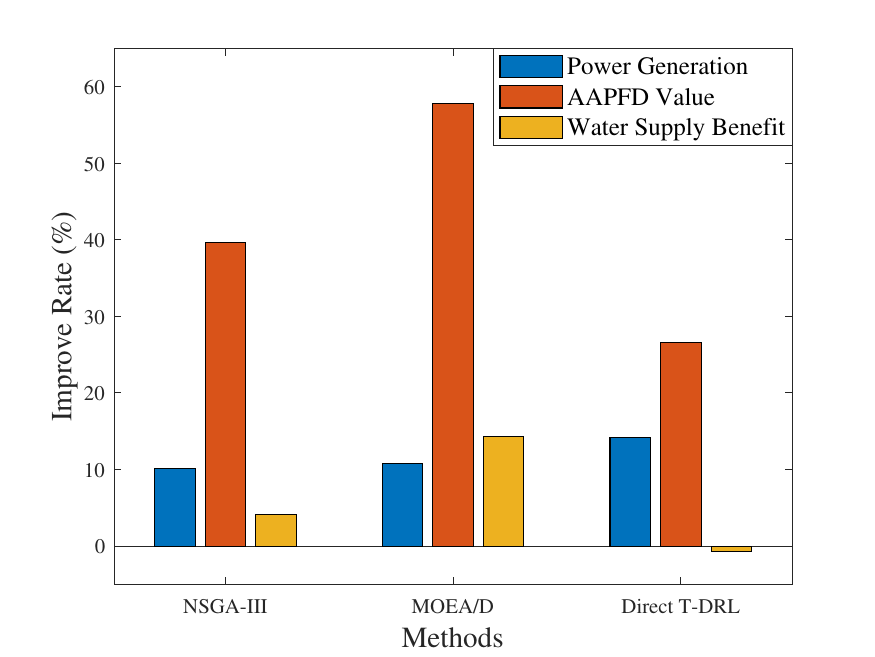}
	\caption{\textrm{Improvements of the three objectives by adopting the Two-stage T-DRL }}
	\label{bar}
\end{figure}

\begin{figure}[ht]
	\centering
	\includegraphics[scale=0.6]{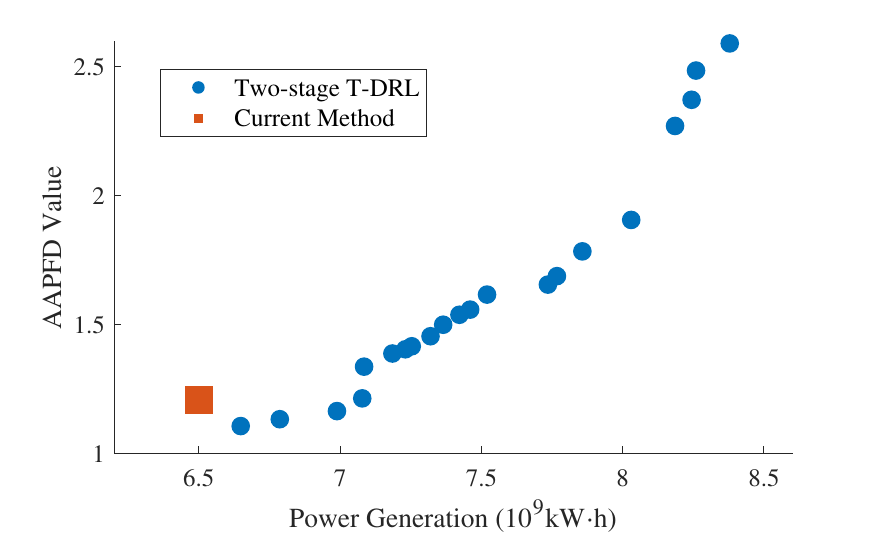}
	\caption{\textrm{Comparison of the current method with Two-stage T-DRL on AAPFD value and power genearation}}
	\label{current}
\end{figure}

\begin{figure}[ht]
	\centering
	\includegraphics[scale=0.55]{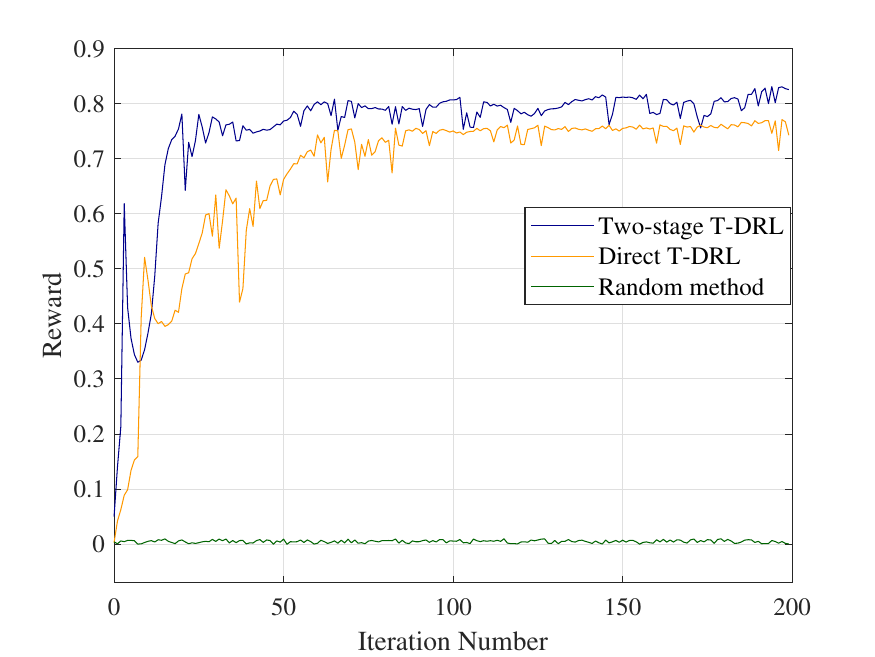}
	\caption{\textrm{Reward for Two-stage T-DRL and Direcr T-DRL under the subproblem of weight combination [0.5,0.25,0.25]}}
	\label{pic:reward}
\end{figure}

\begin{figure}[ht]
	\centering
	\includegraphics[scale=0.55]{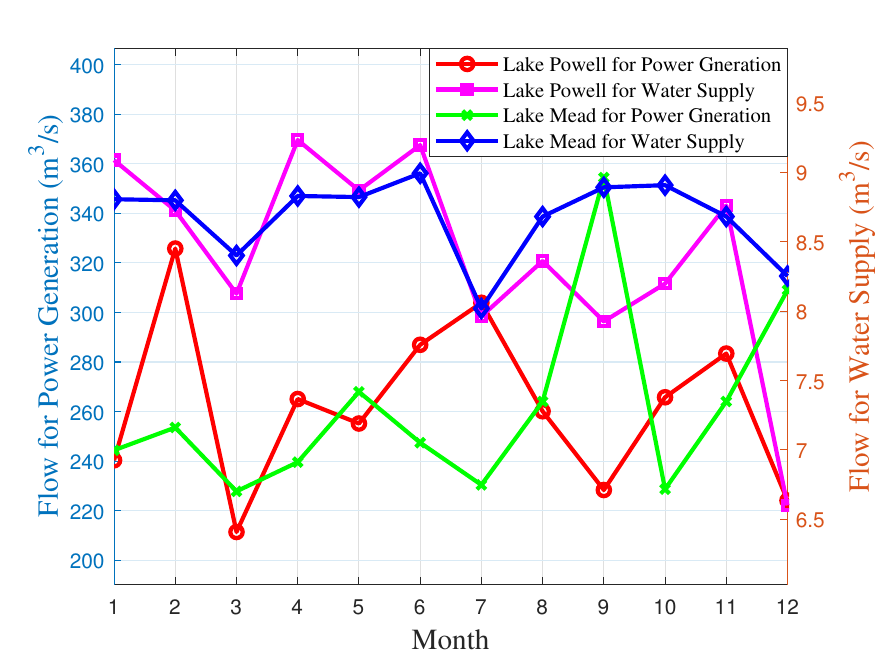}
	\caption{\textrm{The detailed operation scheme generated by Two-stage T-DRL }}
	\label{pic:operation}
\end{figure}

Based on the river data over the past ten years, the most suitable ecological outflows for Lake Mead and Lake Powell are calculated with the annual distribution method, as shown in Table \ref{MSER}. By determining the most suitable ecological outflow, we can categorize the operation months accordingly. Both reservoirs experience a wet season in April and between August and October, and the most suitable ecological outflow during the other months of the year is comparatively low.

\subsection{Results and discussion}\label{sec:sec53}
~\
\subsubsection{Pareto front of the proposed method}\label{sec:sec531}

The Pareto fronts obtained by the Direct T-DRL, Two-stage T-DRL, NSGA-III, and MOEA/D methods are displayed in Figure \ref{PF}. As illustrated by the three-dimensional Pareto frontier in Figure \ref{PF1}, it is evident that the proposed Two-stage T-DRL method outperforms the other two evolutionary algorithms and Direct T-DRL. This superior performance can be attributed to the fact that, in T-DRL, each Pareto-optimal solution in the Pareto front represents a weight combination, with T-DRL consistently focused on solving the single-objective optimization model for this set of weight combinations. In contrast, multiobjective evolutionary algorithms often employ nondominated sorting techniques, resulting in Pareto-optimal solutions that are not guaranteed to be optimal.

Moreover, the two-stage embedding progress enhances the ability of the T-DRL method to effectively extract and learn information. Additionally, the performance of evolutionary algorithms is heavily reliant on the quality of the initial population. Moreover, the T-DRL method utilizes a neural network which has been extensively researched, and parameters can be appropriately adjusted to obtain a satisfactory solution.

Figure \ref{PF2} displays the Pareto fronts as viewed from the X and Z axes. The majority of Pareto-optimal solutions generated by the Two-stage T-DRL method are superior to those produced by the other three methods. Notably, an increase in the value of objective function 1 results in a reduction in the value of objective function 3. Figure \ref{PF4} depicts the Pareto frontier from the perspective of the X and Y axes, where an increase in the value of objective function 1 corresponds to an increase in the value of objective function 2.

Examining the Pareto fronts from four angles reveals that all the T-DRL methods perform better than the evolutionary algorithms in terms of objective function 2 and objective function 3.
This is because the random crossover positions of the chromosomes and the random mutation positions influence the results of the evolutionary algorithms. Given that the problem involves multiple binary variables and continuous variables, the evolutionary algorithms struggle to obtain good solutions compared to the learning strategy employed in T-DRL methods.

All Pareto-optimal solutions obtained with the proposed Two-stage T-DRL method are compared with those produced by the NSGA-III method and the Direct T-DRL method. Compared to the NSGA-III method, the Two-stage T-DRL method provides a solution that involves generating 10.11\% more electricity, reducing the amended annual proportional flow deviation by 39.69\%, and increasing the water supply revenue by 4.10\%. In comparison to the Direct T-DRL method, the Two-stage T-DRL method provides a solution that involves generating 14.1852\% more electricity and reducing the amended annual proportional flow deviation by 26.5454\%. Figure \ref{bar} illustrates the superior performance of the proposed method compared to other methods across all three objective functions.

\subsubsection{Comparison with the current method} \label{sec:sec532}
To demonstrate the feasibility and superiority of the proposed Two-stage T-DRL method, we compare it with the actual operation strategies of the two hydropower reservoirs. The data for these strategies were sourced from the U.S. Bureau of Reclamation website \cite{usbr_data}.

The existing reservoir operation method is primarily focused on power generation. We transform the three-objective scheduling optimization problem into a biobjective operation optimization problem involving power generation and ecological protection. The results of our method and the current operation scheme are displayed in Figure \ref{current}. The results of the current method are inferior to those of Two-stage T-DRL. Consequently, when compared to the current practices at Lake Mead and Lake Powell, the Two-stage T-DRL method yields better operational outcomes.

\subsubsection{Performance analysis of the proposed scheme}\label{sec:sec533}
We compare the performance of Two-stage T-DRL and Direct T-DRL based on a subproblem of the MMROO problem with a weight combination of [0.5, 0.25, 0.25]. Figure \ref{pic:reward} shows the rewards at various iterations for this weight combination; the blue line represents the change in the reward obtained with the Two-stage T-DRL method, the orange line represents the result of the Direct T-DRL method, and the green line represents the results of a random method. It is apparent that the T-DRL method with two-stage embedding progress exhibits a faster convergence speed and better performance than T-DRL method with direct embedding progress.

For this particular subproblem, the detailed operation scheme generated by the Two-stage T-DRL method is illustrated in Figure \ref{pic:operation}.

\section{Conclusions}\label{sec:sec6}
~\
In this paper, we investigate a multiobjective multihydropower reservoir joint operation strategy in which power generation, environmental protection, and water supply are concurrently optimized. The substantial decision variable space, comprising continuous and binary variables, coupled with the numerous constraints in reservoir operation, pose significant challenges. To tackle this problem, a transformer-based deep reinforcement learning method is established to train the model to efficiently and automatically solve the multiobjective optimization problem. Moreover, we propose a two-stage embedding progress in the encoder progress to better learn the information. Our experimental results reveal that the T-DRL method with two-stage embedding progress demonstrates superior information extraction capabilities compared to a T-DRL method with direct embedding progress. Moreover, when compared to evolutionary algorithms, the T-DRL method exhibits enhanced performance in solving problems with binary decision variables. Additionally, the T-DRL method, through its decomposition strategy, showcases a more extensive ability to search for solutions than do the existing evolutionary algorithms.
\section*{Acknowledgement}
This work is supported by the National Natural Science Foundation of China under Grant 62171218.

\bibliographystyle{unsrt}
\bibliography{refs}
\end{document}